\documentclass[letterpaper, 10 pt, conference]{ieeeconf}  

\IEEEoverridecommandlockouts                              

\usepackage{graphics} 
\usepackage{epsfig} 
\usepackage{amsmath} 
\usepackage{amssymb}  
\usepackage{caption}
\usepackage{subcaption}
\usepackage{booktabs}
\usepackage[colorlinks=true, allcolors=black]{hyperref}

\title{\LARGE \bf APR-Transformer: Initial Pose Estimation for Localization in Complex Environments through Absolute Pose Regression}

\author{Srinivas Ravuri$^{1}$, Yuan Xu$^{1}$, Martin Ludwig Zehetner$^{2}$, Ketan Motlag$^{1}$, and Sahin Albayrak$^{1}$
\thanks{$^{1}$Faculty of Electrical Engineering and Computer Science, Chair of Agent Technology, Technische Universität Berlin, Straße des 17. Juni 135, 10623 Berlin, Germany
        {\tt\small (srinivas.ravuri@dai-labor.de)}}%
\thanks{$^{2}$This research was conducted at Faculty of Electrical Engineering and Computer Science, Technische Universität Berlin, Berlin, Germany {\tt\small (zehetner@fzi.de)}}
}

\begin{document}

\maketitle
\thispagestyle{empty}
\pagestyle{empty}

\begin{abstract}

Precise initialization plays a critical role in the performance of localization algorithms, especially in the context of robotics, autonomous driving, and computer vision. Poor localization accuracy is often a consequence of inaccurate initial poses, particularly noticeable in GNSS-denied environments where GPS signals are primarily relied upon for initialization. Recent advances in leveraging deep neural networks for pose regression have led to significant improvements in both accuracy and robustness, especially in estimating complex spatial relationships and orientations. In this paper, we introduce APR-Transformer, a model architecture inspired by state-of-the-art methods, which predicts absolute pose (3D position and 3D orientation) using either image or LiDAR data. We demonstrate that our proposed method achieves state-of-the-art performance on established benchmark datasets such as the Radar Oxford Robot-Car and DeepLoc datasets. Furthermore, we extend our experiments to include our custom complex APR-BeIntelli dataset. Additionally, we validate the reliability of our approach in GNSS-denied environments by deploying the model in real-time on an autonomous test vehicle. This showcases the practical feasibility and effectiveness of our approach. The source code is available at: \url{https://github.com/GT-ARC/APR-Transformer}.

\end{abstract}

\section{Introduction}

In recent years, there has been an uptrend in the utilization of image data for vision-based localization, employing diverse methodologies such as image retrieval \cite{noh2018largescale, arandjelović2016netvlad, Ali-bey_Chaib-draa_Giguère_2023, keetha2023anyloc}, feature matching \cite{panek2022meshloc, Yang_2022_CVPR}, scene coordinate regression \cite{10044413, fang2022accurate}, and pose regression \cite{kendall_posenet_2015, brahmbhatt_geometry-aware_2018, valada18icra, radwan_vlocnet_2018, shavit_learning_2021, 10.1007/978-3-031-20080-9_9, wang_atloc_2020, wang2023robustloc}. These techniques can, for example, be used in the re-localization or initialization of an autonomous vehicle in a GNSS-denied environment. Contrary to the image retrieval approaches which require large memory databases, pose regression predicts poses directly from a given image. Pose regression is aimed at estimating the six degrees of freedom (6-DoF) pose of an autonomous agent using sensors such as LiDARs, cameras, and radars. This capability is pivotal in various applications such as autonomous driving, robot navigation, and augmented reality. Conventional approaches for estimating position and orientation present various trade-offs, with geometric methods often struggling in complex environments. In contrast, deep learning-based pose regression methods have demonstrated notable performance across diverse environments, offering significant potential to improve localization accuracy and robustness. 

\begin{figure}[ht!]
\centering
\includegraphics[width=0.5\textwidth]{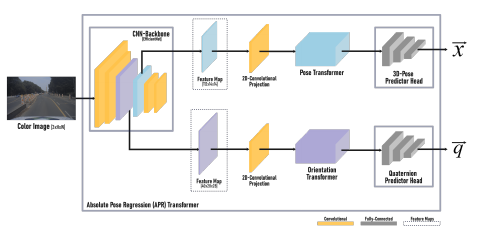}
\caption{APR-Transformer model architecture employs absolute pose regression with two Transformers separately querying features for position and orientation. This architecture utilizes a single camera image as input; features from a CNN backbone (EfficientNet variants) are extracted which are further processed by the Transformers and MLP heads regressing position and orientation.}\label{APR-Transformer-image}
\end{figure}

Despite the promising results demonstrated by vision-based localization approaches, they remain susceptible to changing weather conditions. LiDAR data on the other hand is more robust towards such a changing environment and provides a depth information which could be leveraged for precise re-localization. Similar to image-based approaches LiDAR data can also be used for retrieval as in \cite{8578568}, but proves to be computationally costly. In contrast, LiDAR-based pose regression \cite{wang_hypliloc_2023} can prove to be computation and storage cost-efficient.  

In this work, we propose a deep-learning model APR-Transformer, see Figure \ref{APR-Transformer-image}, for Absolute Pose Regression (APR) based on images or LiDAR data from onboard cameras and sensors of a vehicle. The multi-pose estimates given by the model during inference are global within a given map. This way, we plan to improve the initialization of the localization algorithm in GNSS-denied environments. Our approach makes the following contributions:

\begin{itemize}
    \item We propose the formulation for multi-camera or LiDAR pose regression leveraging state-of-the-art CNN backbones with a Transformer encoder-decoder network. During training, the model uses multi-camera images or LiDAR point cloud (2D bird's eye view or raw points) as input and then regresses the pose utilizing the ground-truth pose labels that are generated from localization results using the LiDAR-based SLAM \cite{shan2020liosam}. This way, our approach is independent of GNSS. 
     \newline 
    \item  The feasibility of the proposed hypothesis is experimentally demonstrated on open-source benchmark datasets, i.e., the DeepLoc \cite{valada18icra} and Oxford RobotCar \cite{RadarRobotCarDatasetArXiv} datasets. Additionally, a custom dataset captured from the BeIntelli \cite{noauthor_beintelli_nodate} autonomous test vehicle around highly dynamic road segments, which include multiple pedestrian crossings, road intersections, and roundabouts is presented. The regressed pose can be used in real-time as an initial guess by any localization algorithm such as NDT scan matching \cite{aoki_error_2023}.   
\end{itemize}

\section{Related Work}
In this section, we present an overview of the current research landscape regarding image and LiDAR-based pose regression methods. These methods can be broadly classified into  Visual Place Recognition (VPR), Relative Pose Regression (RPR), and Absolute Pose Regression (APR) based on the task and inputs utilized during inference and their algorithmic attributes.


\subsection{Image-based APR}
Image-based APR involves training deep neural networks to directly predict the 6-DoF pose of a vehicle using images captured by onboard cameras. This involves a localization function which extracts features from a given scene. High-dimensional pose components are constructed from the non-linear embedding of these extracted features. Finally, poses are learned by a network of these components. This is typically achieved through supervised learning. Usually, a CNN-based backbone network is used to extract the features, which are then embedded in a high-dimensional space. This embedding is used to regress the camera pose in the scene. This approach of pose regression is presented in PoseNet \cite{kendall_posenet_2015} and its variants \cite{brahmbhatt_geometry-aware_2018, walch_image-based_2017,  naseer_deep_2017}. Many other approaches \cite{radwan_vlocnet_2018, shavit_learning_2021, wang_atloc_2020, zhu_map-net_2021} differ mainly in the network architecture and loss function. Annotated datasets such as \cite{valada18icra} or \cite{RadarRobotCarDatasetArXiv} containing pairs of images and their corresponding ground truth poses are used for training.

\subsection{LiDAR-based APR}
LiDAR-based APR involves utilizing point cloud data captured by LiDAR sensors to directly estimate the 6-DoF pose of a vehicle. Contrary to camera-based APR, LiDAR sensors are less affected by variations in lighting conditions compared to cameras, making them suitable for pose estimation in challenging lighting environments, such as low-light conditions or nighttime driving. LiDAR sensors provide rich 3D spatial information, allowing for more accurate depth estimation and obstacle detection compared to cameras. This 3D perception can enhance the accuracy and reliability of absolute pose regression models, especially in scenarios where precise spatial understanding is crucial, such as navigating complex urban environments or avoiding obstacles. Given such highly informative LiDAR data, it is worth exploring the possibility of directly regressing pose from LiDAR output. This can be in either a raw point cloud, a bird's eye view, or a flattened image from a scan with its respective pose. Similar to image-based approaches, deep learning techniques such as \cite{wang_hypliloc_2023} use the spherical projection image often referred to as signal/range image, constructed from a LiDAR scan along with its respective pose as a training input. 

\begin{figure*}[!t]
    \centering
    \includegraphics[width=\linewidth]{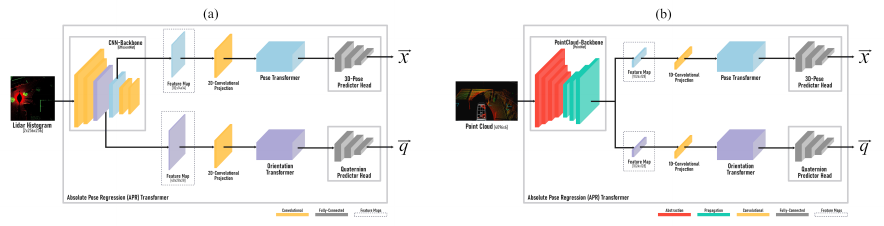}
    \label{APR-Transformer-lidar-fig}
    \caption{We experiment with different architecture design choices and input features by converting LiDAR data to task-specific modalities. a) 2D histogram representation of input LiDAR data; EfficientNet is used as the backbone to extract multi-resolution features further processed by the Transformer blocks performing pose regression. b) PointNet++ is used to extract features from raw 3D points of LiDAR data further processed by the Transformer blocks performing pose regression.}
    \label{apr-transformer lidar modality}
\end{figure*}



\section{Learning Ego-Pose Regression}
The ego-pose $p$ of an agent can be represented as a tuple ⟨$x$,$q$⟩, wherein $x$ denotes the 3D-position of the agent and $q$ is the quaternion encoding of the 3D-orientation. Following recent wide-spread use of Transformers and their successful application in various computer vision tasks \cite{Ali-bey_Chaib-draa_Giguère_2023, carion2020endtoend, dosovitskiy2021image}, we propose a modality-adaptive model architecture leveraging separate position and orientation Transformer encoders for robust aggregation of feature maps or vectors computed by modality-dependent backbones \cite{shavit_learning_2021}. To this end, the following section details modality-specific preprocessing steps and variations of the proposed APR-Transformer models, as well as the adaptive pose regression loss \cite{kendall2017geometric} function used. 

\subsection{Modality-Specific Data Pre-Processing} \label{modal specific tasks}
We propose performing pose regression utilizing different modality types, i.e., images, LiDAR-based histograms, and point cloud representations, with input pre-processing steps tailored to each modality to ensure effective feature extraction and representation.

\paragraph{Image-based}\label{image based}
As an initial step, raw input images are pre-processed by scaling them to a resolution of $256 \times 256$ pixels. Afterwards, we normalize the images by pixel mean subtraction and standard deviation division of the ImageNet dataset. Additionally, the target variables are then Min-Max normalized. Furthermore, during training, due to improved observed performance on cross-weather datasets, we randomly perform augmentations by perturbing the brightness, saturation, contrast, and hue of the images and also perform weather augmentations using the albumentations library \cite{2018arXiv180906839B}. The architecture of the APR-Transformer model, designed to process the so pre-processed images, is shown in Figure \ref{APR-Transformer-image}.

\paragraph{LiDAR-based}\label{lidar based}
We propose two distinct strategies to utilize LiDAR-based point cloud data in the context of the APR-Transformer. The corresponding model architectures can be seen in Figure \ref{apr-transformer lidar modality}. In general, when converting LiDAR data to task-specific modalities, it can be observed that each modality incorporates multiple interpretations of neighboring points, enhancing the depth and semantic information available for subsequent analysis and processing. 

Our first strategy follows \cite{chitta2022transfuser}. Accordingly, we project the 3D points into a 2-bin histogram across a 2D bird's-eye-view (BEV) grid with fixed resolution. Focusing on a region extending 30 meters in front of the ego-vehicle and 50 meters to each side, we cover a BEV grid measuring $32 \times 32$ meters resulting in a resolution of $256 \times 256$ pixels.  

In our second strategy, we reduce the number of points considered to $4096$ to accommodate our model backbone requirements \cite{qi2017pointnet} by first cropping the raw points of the LiDAR frame to include only points within a 20 meter radius of the agent, and then we subsample the raw point cloud data using farthest point sampling.  Additionally, we normalize the coordinates of the considered points and concatenate the absolute and relative coordinates. 

\subsection{APR-Transformer}

Given a LiDAR point cloud scan $\textbf{L}\in\mathbb{R}^{N\times C}$ or image $\textbf{I}\in\mathbb{R}^{H\times W\times C}$, the rough structure of our APR-Transformer architecture can be described as follows. The input modality selected is fed into a backbone that extracts two feature maps or vectors. Each of the two feature representations is used by one subsequent model branch. The first branch computes the prediction of the agent's 3D position, while the second branch estimates the quaternion encoding of the 3D orientation. The two branches share an identical structure, in which the feature representations are first projected using a convolutional model component, then fed into a Transformer, and finally, a fully connected MLP model head with one hidden layer followed by a ReLU activation is used to regress the target vector and compute the pose predictions. 

We use the standard Transformer Encoder-Decoder \cite{carion2020endtoend} architecture with $L$ identical layers and perform Layer Normalization (LN) before each multi-head attention module and residual connection. The formulations of the multi-head attention in Transformer Encoder-Decoder architecture are omitted for simplicity reasons, we refer the reader to \cite{vaswani2023attention}. 

\paragraph{Image \& Lidar-based Histogram}\label{image and lidar historam} When considering images and LiDAR-based BEV histograms, we use an off-the-shelf state-of-the-art pre-trained convolutional backbone to extract feature maps at multiple resolutions. We follow the same procedure as in \cite{carion2020endtoend} for creating embeddings for Transformer-compatible inputs. Furthermore, each position in the feature map is assigned with a learned encoding to preserve the spatial information of each location.  

We experiment with different EfficientNet \cite{tan2020efficientnet} variants and employed transfer learning to utilize the weights of the models pre-trained on the ImageNet dataset for the image and LiDAR-based BEV histogram modalities. As per \cite{shavit_learning_2021} we sample the feature maps at two different endpoints. For the selected EfficientNet variants we extract feature maps $F_{x}$ and $F_{q}$ at resolutions (\textit{N}, 112, 14, 14) and (\textit{N}, 40, 28, 28), with \textit{N} being the batch size. 

\paragraph{Lidar-based Points}\label{lidar points} 

When leveraging LiDAR-based point data, we employ a PointNet++ \cite{qi2017pointnet} based backbone, which we initialize by pre-training a complete PointNet++ model, consisting of four abstraction layers, four propagation layers and a classification head, on the SemanticKITTI \cite{behley2019iccv} dataset for semantic segmentation. Afterwards, we remove the last propagation layer and classification head and use the remaining components as backbone for our APR-Transformer. We utilize the output of the last remaining propagation layer as feature vectors $F_{x}$ and $F_{q}$ at resolutions of (\textit{N}, 128, 1024). With each of the 128 vectors corresponding to a reduced point of the original 4096-point data input. 

The Transformer-compatible input embeddings and associated learned encodings, preserving the spatial information of the backbone outputs, are then computed by first separating the 128 vectors into eight groups based on the absolute $z$-coordinates, i.e., height, of their corresponding reduced points. Subsequently, the 16 feature vectors per group are sorted in a $4 \times 4$ grid based on the $x$ and $ y$ coordinates of their reduced points. Afterward, we adapt the procedure used in the image-based APR-Transformer case by computing the learned positional encodings along the three axes to generate the final Transformer inputs.




\subsection{Pose Regression and Loss Function}

\begin{align}
   \label{positionloss}
   L_{p}(x) = \sum_{i=1}^{D}|x_{p_i}-x_{t_i}|  \\
   \label{orientationloss}
   L_{o}(q) = \sum_{i=1}^{D}|q_{p_i}-q_{t_i}| 
\end{align} 
\begin{align}
  \label{loss}
  L_{pose} = L_{p}\exp(-s_{x}) + s_{x} +  L_{o}\exp(-s_{q}) + s_{q}
\end{align} 

We train the model variants to minimize the position loss $L_{p}$, see \autoref{positionloss}, and orientation loss $L_{o}$, see \autoref{orientationloss}, for the ground truth pose, where $L_{p}$ and $L_{o}$ are $L_1$ losses. We combine the position and orientation losses using the formulation by Kendall et al. \cite{kendall2017geometric} shown in \autoref{loss}. Where $s_{x}$ and $s_{q}$ are learned parameters that control the balance between the position loss and the orientation loss.

\section{Experiments}
To study the effect of different architecture design choices and input feature types, we conducted multiple ablation experiments, focusing primarily on outdoor datasets collected from vehicle agent setups. Consequently, in this section, we evaluate the APR-Transformer on two established open-source benchmarks, i.e., the DeepLoc \cite{valada18icra} and Radar Oxford Robot-car \cite{RadarRobotCarDatasetArXiv} datasets. Further, we compare the performance of the APR-Transformer against state-of-the-art methods and demonstrate its practical application in real-time scenarios. Additionally, we extend our experiments to the custom APR-BeIntelli dataset with more dynamic and challenging scenes in a dense city environment.

\subsection{Datasets}
\paragraph{DeepLoc}
A large-scale urban outdoor localization dataset was collected using an autonomous robot platform. RGB images were captured at a resolution of $1280 \times 720$ pixels at 20 Hz. The dataset is currently comprised of only images and one scene spanning an area of $110 \times 130$ meters,  with different driving patterns. The ground truth pose labels are computed using a LiDAR-based SLAM system with sub-centimeter and sub-degree accuracy \cite{valada18icra}.

\paragraph{The Radar Oxford Robot-Car}
A large-scale multi-modal outdoor dataset \cite{RadarRobotCarDatasetArXiv} collected while driving an autonomous vehicle through Oxford, UK, was captured over a year with many combinations of weather, traffic, and pedestrians. The dataset offers LiDARs, Camera, GNSS, and Radar information enabling a thorough evaluation of our models. We train the same configuration of the models utilizing the camera images and LiDAR point clouds respectively as explained in section \ref{modal specific tasks}.  We use the same dataset split and report the average median position and orientation error as per the benchmark \cite{RadarRobotCarDatasetArXiv, RobotCarDatasetIJRR}. 

\paragraph{APR-BeIntelli}
To study the practical applicability of APR-Transformer, we scale our experiments to a custom dataset created as a part of the ongoing research project BeIntelli \cite{noauthor_beintelli_nodate} led by the Technische Universität Berlin, DAI Labor, and perform multiple ablation studies. We name the dataset as APR-BeIntelli. The dataset was collected using our autonomous test vehicle equipped with a multi-sensor setup (Cameras, LiDARs, Radars, Ultrasonics). Multi-camera RGB images (front-center, front-left, and front-right) were captured at a resolution of $1208 \times 1920$ pixels, with synchronized LiDAR point clouds, GPS, IMU, and other sensory information across our BeIntelli test track in Berlin across multiple days with changing weather conditions. The dataset can be used for vision-based applications such as global localization, camera re-localization, object detection, semantic segmentation, etc. Lighting, weather conditions, and temporary construction zones make the dataset challenging for vision-based localization tasks. To generate the ground truth pose labels for the pose regression task, we leverage the LiDAR-based SLAM \cite{shan2020liosam} techniques and use the computed localization pose as ground truth similar to the DeepLoc dataset. The dataset was generated with minimal effort, as we leveraged the same data collection run used for generating the LiDAR-based point cloud map, incorporating additional camera data. This streamlined approach significantly reduced the resource-intensive data acquisition process while maintaining data integrity. The dataset is divided into a train and a test split collected at different times and weather conditions. The train set contains 3437 frames of multi-camera images and point clouds with multiple loops while the test set consists of 902 frames from a different sequence of collection. We aim to further expand the dataset by incorporating additional sensor modalities and diverse sequences captured under varying weather conditions and intervals. We intend to publish the dataset along with the accompanying source code.

\subsection{Experiment Setup}\label{experiment-setup}
The APR-Transformer model is optimized to minimize the loss function in \autoref{loss}. We employ an Adam optimizer with a step learning rate scheduler and use L=6 encoder and
decoder layers for both the position and orientation Transformers. We experiment with batch sizes of 8 and 16, an initial learning rate $\lambda$ = $10^{-4}$ and weight decay of $5^{-4}$. The model is trained using PyTorch \cite{paszke2019pytorch} for 300 epochs and evaluated on the test subset of the respective datasets. Experiments reported were performed on a 24GB NVIDIA RTX 3090 GPU. 

\subsection{Results on DeepLoc dataset}
APR-Transformer was optimized on the DeepLoc training set which consists solely of camera images from seven distinct driving loops, with alternating driving styles amounting to 2737 images. The model was inferred on the test set comprising 1174 images and the performance of the model is depicted in \autoref{fig:deeploc_results} with median pose error (0.7m) and median rotation error (3.354deg).

\subsection{Results on Radar Oxford Robot-Car dataset}
\begin{figure*}[!t]
    \centering
    \begin{minipage}[b]{0.3\linewidth}
        \centering
        \includegraphics[width=\linewidth]{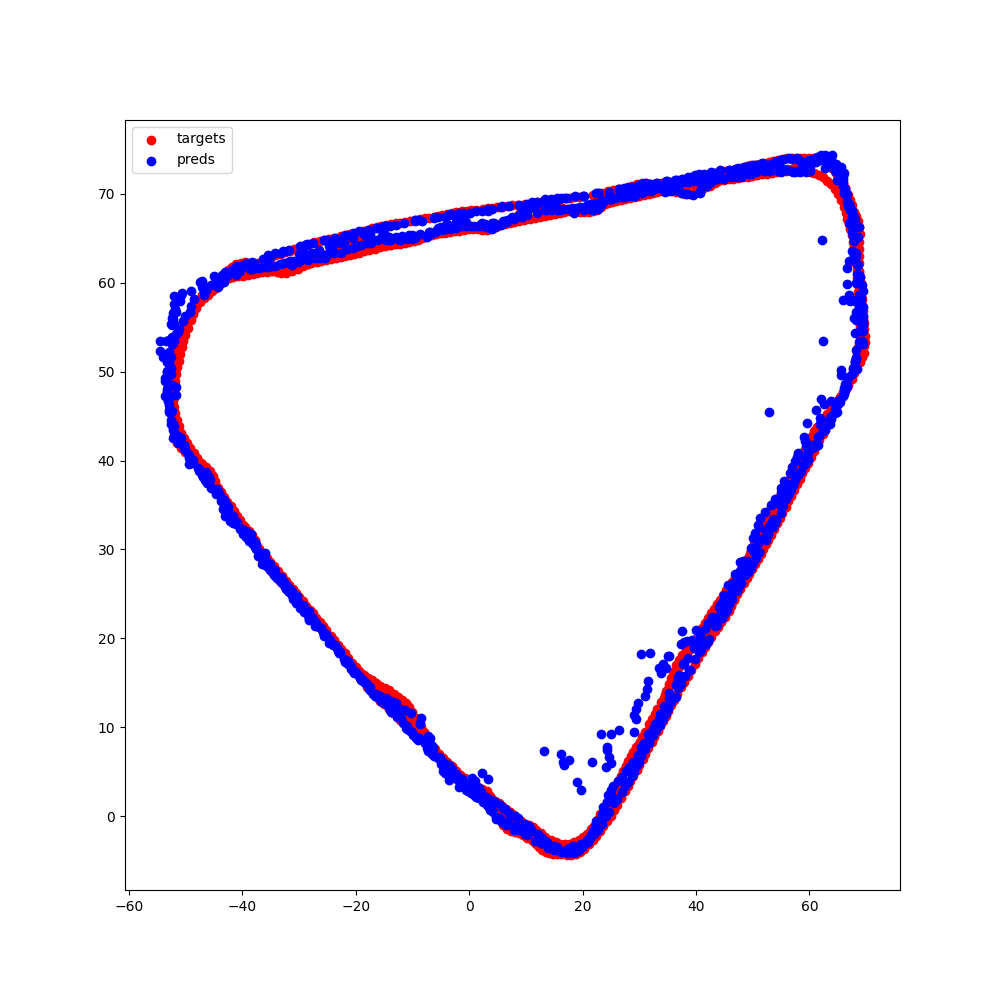}
        \subcaption{DeepLoc dataset (0.7m, 3.35\textdegree)}
        \label{fig:deeploc_results}
    \end{minipage}
    \hfill
    \begin{minipage}[b]{0.3\linewidth}
        \centering
        \includegraphics[width=\linewidth]{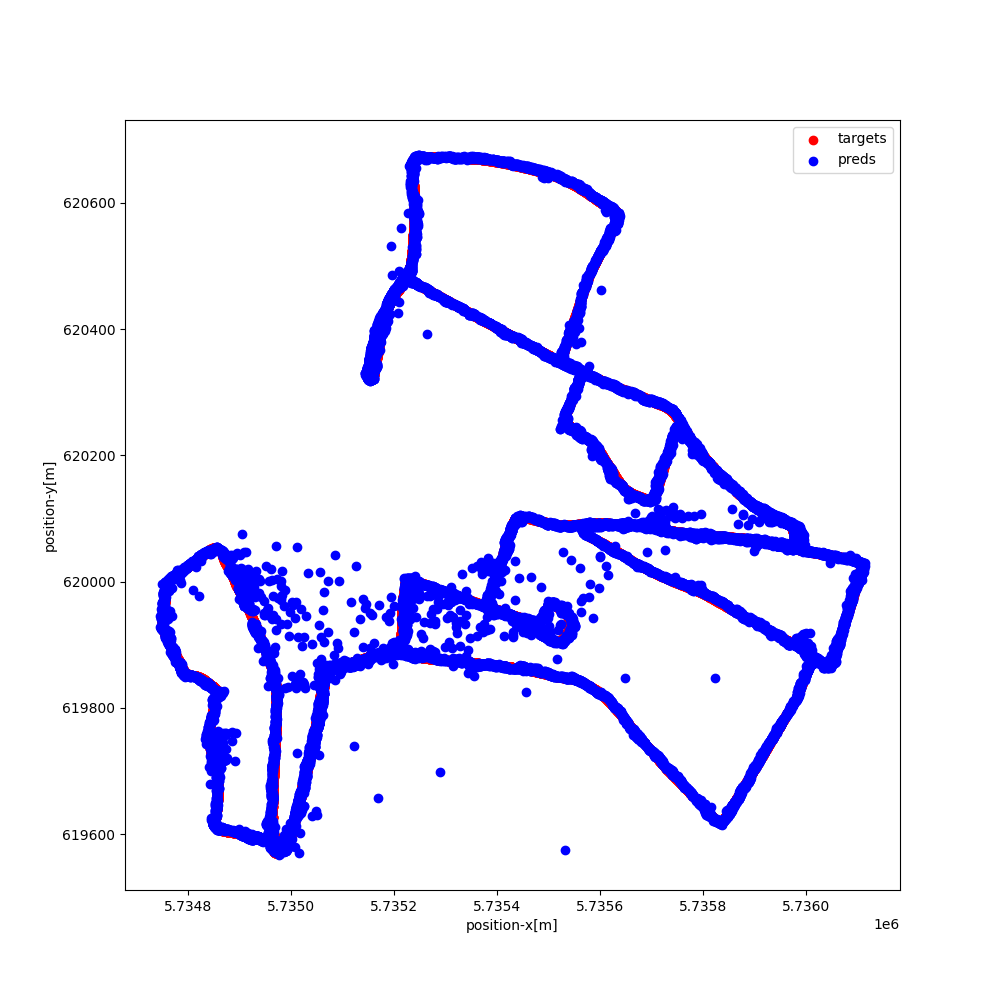}
        \subcaption{Robot-Car (3.65m, 0.60\textdegree)}
        \label{fig:oxford_image_results}
    \end{minipage}
    \hfill
    \begin{minipage}[b]{0.3\linewidth}
        \centering
        \includegraphics[width=\linewidth]{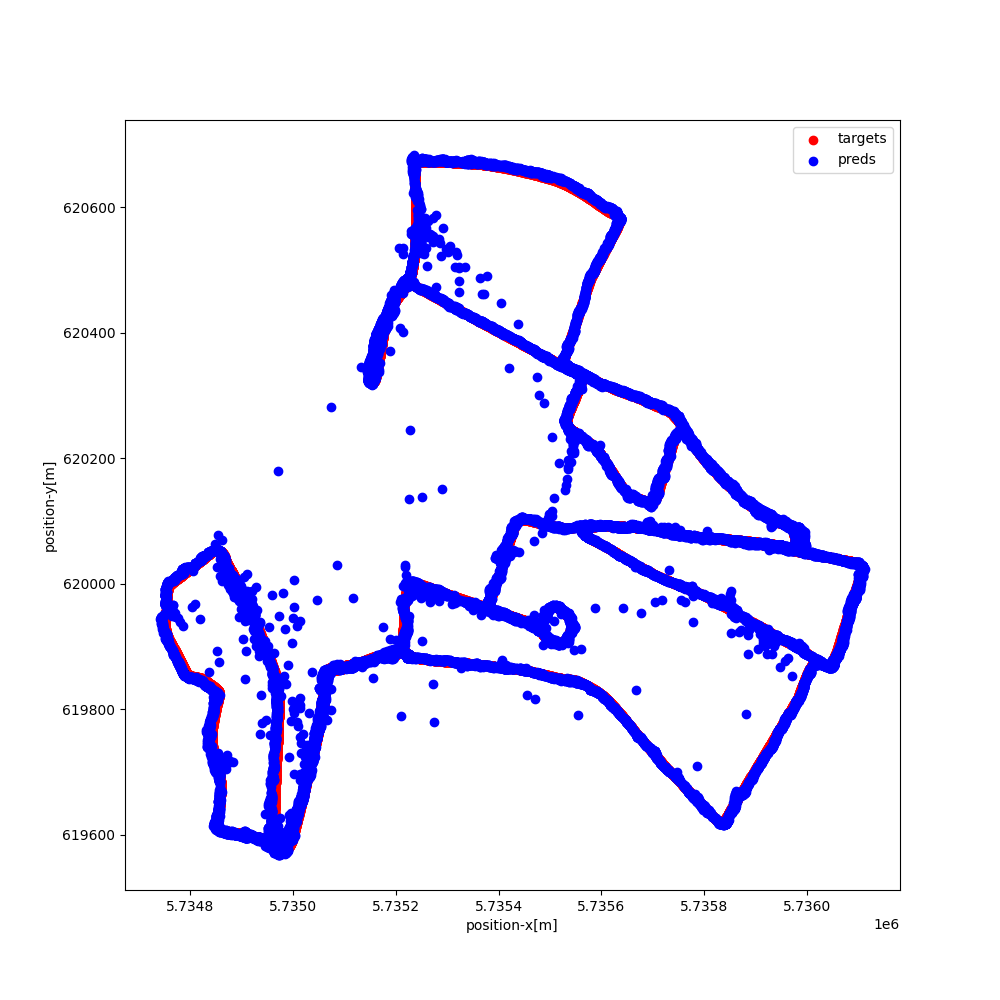}
        \subcaption{Robot-Car (4.13m, 0.63\textdegree)}
        \label{fig:oxford_lidar_results}
    \end{minipage}
    \caption{Trajectory visualizations of APR-Transformer model predictions on respective datasets, with the x and y axes indicating the position in meters. The predicted pose is shown in blue and the ground truth pose in red. Figure \ref{fig:oxford_image_results} shows the performance of the model with 3D LiDAR points as input and Figure \ref{fig:oxford_lidar_results} shows the performance of the model with 2D LiDAR BEV as input.}
    \label{apr-transformer robotcar predictions}
\end{figure*}

APR-Transformer is trained as described in \autoref{experiment-setup} on FULL and LOOP sequences, the model achieves a median pose error of (3.34m) and an orientation error of (1.04deg) on the LOOP route using only camera data. For the FULL routes utilizing both LiDAR modalities, we report the average median pose and average median orientation errors. With 2D LiDAR BEV input and EfficientNetB0 as the backbone, the model achieves an average median position error of (4.85m) and an average median orientation error of (0.60deg). When using raw 3D points as input to the PointNet++ backbone, the model achieves an average median position error of (4.25m) and an average median orientation error of (0.53deg). The experiment results including all the routes are tabulated in Table \ref{tab:radar-robot-car}, with the EfficientNet-B0 variant as the backbone, and the performance of the model is depicted in Figures \ref{fig:oxford_image_results} and \ref{fig:oxford_lidar_results}.

\subsection{Results on APR-BeIntelli dataset}
\begin{figure*}[!t]
    \centering
    \begin{minipage}[b]{0.3\linewidth}
        \centering
        \includegraphics[width=\linewidth]{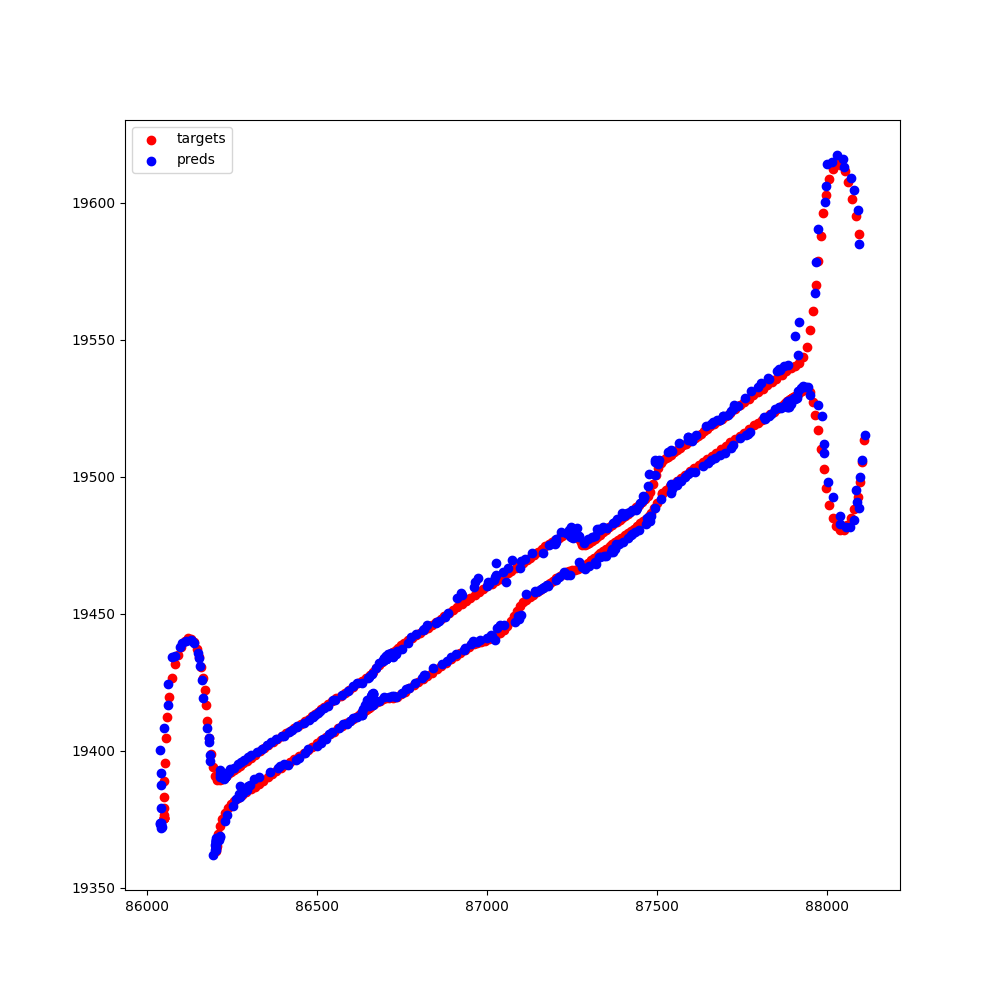}
        \subcaption{Image only as input (7.40m, 0.70\textdegree)}
        \label{fig:beintelli-image-results}
    \end{minipage}
    \hfill
    \begin{minipage}[b]{0.3\linewidth}
        \centering
        \includegraphics[width=\linewidth]{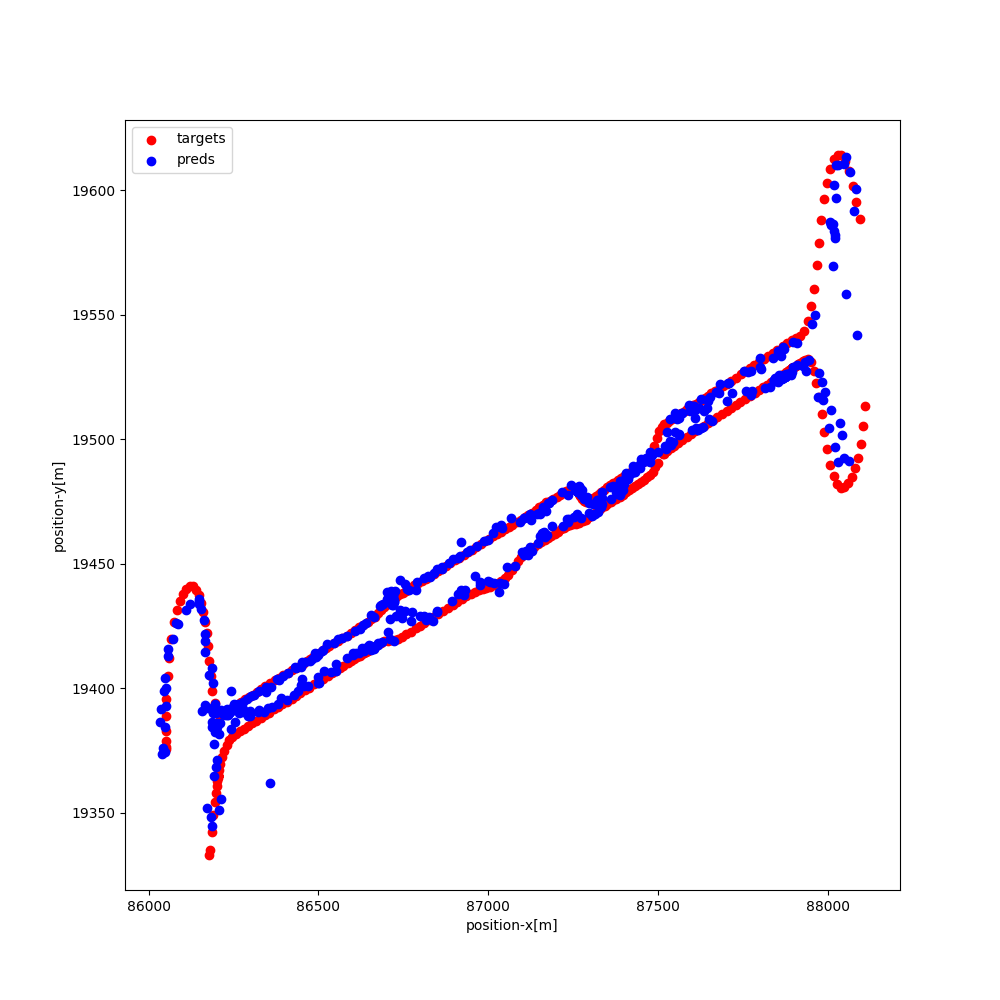}
        \subcaption{2D histogram (8.48m, 1.42\textdegree)}
        \label{fig:beintelli-histogram-results}
    \end{minipage}
    \hfill
    \begin{minipage}[b]{0.3\linewidth}
        \centering
        \includegraphics[width=\linewidth]{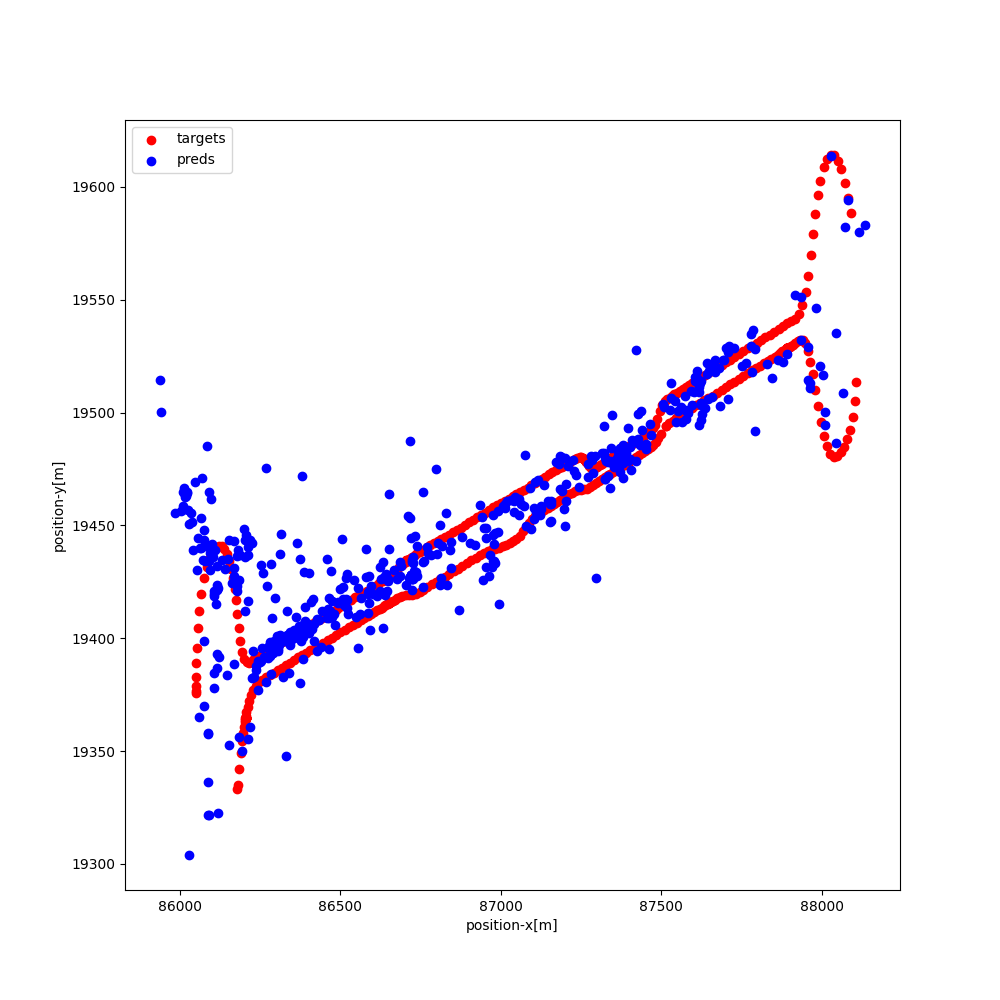}
        \subcaption{3D points (126.75m, 12.25\textdegree)}
        \label{fig:beintelli-points-results}
    \end{minipage}
    \caption{Trajectory visualizations of APR-Transformer model predictions on the APR-BeIntelli dataset, with the x and y axes indicating the position in meters. The predicted pose is shown in blue and the ground truth pose in red.}
    \label{apr-transformer beintelli predictions}
\end{figure*}

APR-Transformer demonstrates its effectiveness with different input modalities. When utilizing image-only input, the model achieves a median pose error (7.40m) and median orientation error (0.70deg) on the test set of the APR-BeIntelli dataset, as depicted in Figure \ref{fig:beintelli-image-results}. Similarly, with LiDAR data as input, incorporating both 2D histogram and 3D raw points, the model achieves median pose errors (8.48m, 1.42deg) and (126.75m, 12.25deg) illustrated in Figures  \ref{fig:beintelli-histogram-results} and \ref{fig:beintelli-points-results}. Overall, APR-Transformer learns a data-driven representation for pose regression. Comparing the camera and 2D LiDAR BEV baseline models, there is no significant deviation observed in pose and orientation errors as tabulated in Table-\ref{tab:apt-transformer beintelli performance}. The experiments conducted on the DeepLoc and Oxford RobotCar datasets yield promising results, comparable to the state-of-the-art, and even outperforming it in certain scenarios. A detailed comparison of the metrics with the state-of-the-art is provided in the \autoref{results}. Moreover, we attribute the superior performance of the camera and 2D LiDAR BEV baseline models to the inherent semantic information present in their inputs, which is especially beneficial for handling dynamic objects and environmental complexity, compared to the model trained solely on raw 3D LiDAR points.

\begin{figure}[ht!]
    \centering
    \includegraphics[width=0.8\linewidth]{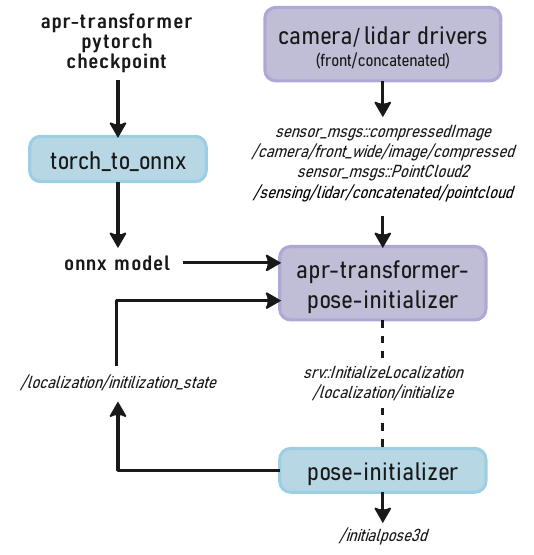}
    \caption{APR-Transformer ROS2 node integrated into Autoware.universe environment.}
    \label{fig:APR-Transformer-ros-diagram}
\end{figure}

\begin{figure}[ht!]
    \centering
    \includegraphics[width=0.8\linewidth]{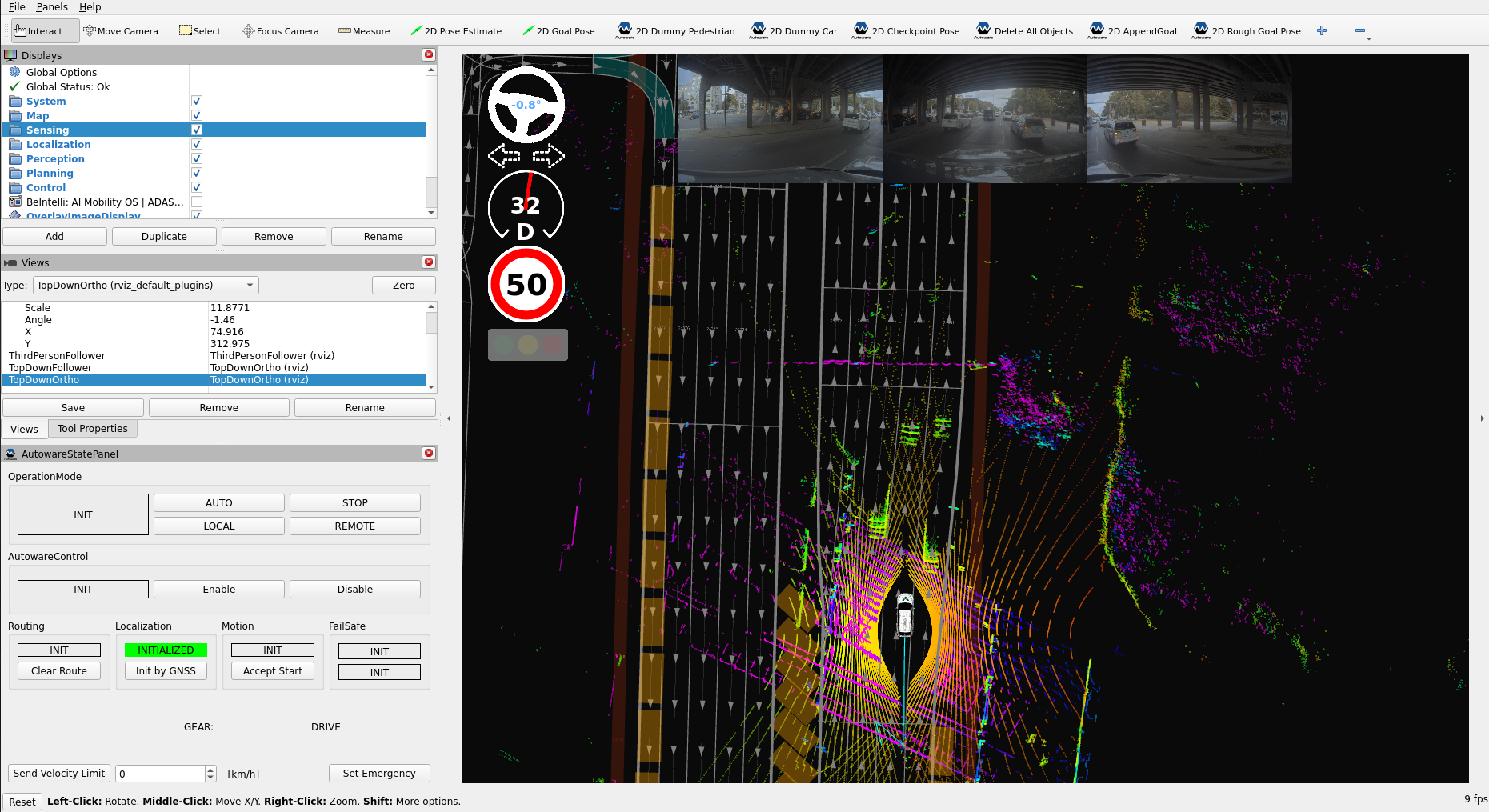}
    \caption{APR-Transformer successfully provided precise pose estimation even in areas beneath a bridge, where GPS signals may be weak. Following successful initialization, the ego vehicle navigates by perceiving its environment.}
    \label{fig:beintelli-initialized}
\end{figure}

\begin{table}[!t]
\caption{Experiment Results of APR-Transformer on the Radar Oxford Robot-Car Dataset. We report the median position/orientation errors in meters/degrees for every test sequence}
\centering
\begin{tabular}{cccc}
\toprule
Modality & Route & Position(m) & Orientation(deg) \\
\midrule
Camera & LOOP & 2.34 & 1.04 \\
2D-LiDAR BEV & FULL-6 & 6.18 & 0.69 \\
2D-LiDAR BEV & FULL-7 & 4.55 & 0.53 \\
2D-LiDAR BEV & FULL-8 & 5.16 & 0.57 \\
2D-LiDAR BEV & FULL-9 & 4.13 & 0.63 \\
3D-LiDAR points & FULL-6 & 5.54 & 0.65 \\
3D-LiDAR points & FULL-7 & 4.50 & 0.53 \\
3D-LiDAR points & FULL-8 & 5.04 & 0.53 \\
3D-LiDAR points & FULL-9 & 3.65 & 0.60 \\
\bottomrule
\end{tabular}
\label{tab:radar-robot-car}
\end{table}

\subsection{Main Results}\label{results}
We conduct a comparative analysis of APR-Transformer against the state-of-the-art models using the DeepLoc and the Radar Oxford Robot-Car datasets. We employ EfficientNet-B0 as the backbone for both the image and LiDAR inputs and report the performance in \autoref{tab:apt-transformer performance}. Our image-based baseline model outperformed VLocNET \cite{valada18icra} on the DeepLoc dataset and is close to state-of-the-art performance on the Radar Robot Car dataset benchmarks considering the simplicity of our model architecture and computational runtime. However, VLocNET++ \cite{radwan_vlocnet_2018} on the DeepLoc dataset achieved a better score, attributed to the auxiliary semantic segmentation task employed by the authors, enabling inductive transfer by harnessing domain-specific information. We believe that through further ablation studies, utilizing a deeper backbone, APR-Transformer can potentially outperform the state-of-the-art RobustLoc \cite{wang2023robustloc} and HypLiLoc \cite{wang_hypliloc_2023}, balancing model performance and inference speed. Nevertheless, the primary focus of this paper is on the model's real-time applicability, enhancing the localization algorithm with an accurate initial estimate, especially in environments with poor GNSS signals.


\section{Integration Into Autoware.Universe Setup}
In the context of the practical applicability of the designed model architecture, APR-Transformer with EfficientNetB0 backbone is integrated into the Autoware.universe software \cite{autoware} stack. We convert the PyTorch model checkpoint into ONNX \cite{bai2019} format and deploy it as a ROS2 node using onnx-runtime. The primary objective is to offer the localization algorithm an alternative source for initial pose estimation in GNSS-denied environments. Here, the model predictions serve as the initial pose estimates for the localization algorithm. An overview of the integration is shown in Figure \ref{fig:APR-Transformer-ros-diagram} and Figure \ref{fig:beintelli-initialized} demonstrates the primary use-case of the model, wherein the predictions serve as an initial pose estimate for the localization algorithm, particularly valuable in urban environments where GPS signal reception is poor, such as under bridges.

\begin{table}[!ht]
\caption{Performance of APR-Transformer on the DeepLoc and Radar Oxford Robot-Car datasets. We report the average median position/orientation errors in meters/degrees. Our results are highlighted in bold against other methods}
\centering
\begin{tabular}{cccc}
\toprule
Method & Dataset & Position(m) & Orient(deg) \\
\midrule
& \multicolumn{2}{c}{Image based} \\
\midrule
 VLocNET\cite{valada18icra} & DeepLoc & 0.68 & 3.43 \\
 VLocNET++\cite{radwan_vlocnet_2018} & DeepLoc & 0.37 & 1.93 \\
 \textbf{APR-Transformer} & \textbf{DeepLoc} & \textbf{0.40} & \textbf{3.35} \\
 MapNet\cite{brahmbhatt_geometry-aware_2018} & Radar RobotCar & 9.84 & 3.96 \\
PoseNet\cite{kendall_posenet_2015} & Radar RobotCar &  25.20 & 17.40 \\
PoseNet+\cite{kendall_posenet_2015} & Radar RobotCar &  55.52 & 11.50 \\
AtLoc\cite{wang_atloc_2020} & Radar RobotCar &  8.54 & 3.54 \\
AtLoc+\cite{wang_atloc_2020} & Radar RobotCar &  5.38 & 1.74 \\
RobustLoc\cite{wang2023robustloc} & Radar RobotCar &  1.97 & 0.84 \\
CoordiNet\cite{moreau2021coordinet} & Radar RobotCar & 2.42& 0.88 \\
\textbf{APR-Transformer} & \textbf{Radar RobotCar} &  \textbf{2.34}&\textbf{ 1.04} \\
 \midrule
 & \multicolumn{2}{c}{LiDAR based} \\
 \midrule
 HypLiLoc\cite{wang_hypliloc_2023} & Radar RobotCar &  3.97 & 0.57 \\
 \textbf{APR-Transformer} & \textbf{Radar RobotCar} &  \textbf{4.63}&\textbf{ 0.63} \\
\bottomrule
\end{tabular}
\label{tab:apt-transformer performance}
\end{table}

\begin{table}[!ht]
\caption{Performance of APR-Transformer on the custom APR-BeIntelli dataset. We report the average median position/orientation errors in meters/degrees}
\centering
\begin{minipage}{\textwidth}
\begin{tabular}{cccc}
\toprule
Method & Modality & Position(m) & Orient(deg) \\
\midrule
& \multicolumn{2}{c}{APR-BeIntelli Dataset} \\
\midrule
 APR-Transformer & Camera & 7.48 & 0.70 \\
 APR-Transformer & 2D-LiDAR BEV & 8.48 & 1.42 \\
 APR-Transformer & 3D LiDAR points & 126.75 & 12.25 \\
\bottomrule
\end{tabular}
\end{minipage}
\label{tab:apt-transformer beintelli performance}
\end{table}

\section{Conclusions}
In this work, we introduced APR-Transformer, a Transformer based encoder-decoder network that learns a data-driven representation for absolute pose regression. Leveraging multi-camera images or LiDAR point clouds (2D bird’s eye view or raw points) as inputs, we assessed the effectiveness of our approach on established benchmark datasets. Our results show that APR-Transformer outperforms many state-of-the-art methods and closely approaches the state-of-the-art in specific scenarios. 

Furthermore, we also focused on the practical applicability of APR-Transformer by utilizing the model's pose estimates in real-time inference as initial poses for the localization algorithm. This enhancement significantly improves initialization as well as localization accuracy, particularly in scenarios with poor GNSS signals. In the future, we aim to extend our methodology by exploring fusion approaches that incorporate additional data modalities yet to be explored in the context of pose regression to the best of our knowledge. Such efforts will contribute to further advancing the field and unlocking new possibilities for improved localization and navigation systems.

\addtolength{\textheight}{-12cm}   


\section*{Acknowledgment}
This work is a part of the BeIntelli project funded by the German Federal Ministry of Transport and Digital Infrastructure under grant agreement number 01MM20004.



\bibliographystyle{ieeetr}
\bibliography{citations}

\end{document}